\documentclass{article}

\usepackage{PRIMEarxiv}

\usepackage[utf8]{inputenc} 
\usepackage[T1]{fontenc}    
\usepackage{hyperref}       
\usepackage{url}            
\usepackage{booktabs}       
\usepackage{amsfonts}       
\usepackage{nicefrac}       
\usepackage{microtype}      
\usepackage{lipsum}
\usepackage{fancyhdr}       
\usepackage{graphicx}       
\graphicspath{{media/}}     

\usepackage{subfig}

\title{A triple-branch network for latent fingerprint enhancement guided by orientation fields and minutiae
}

\author{
  Yurun Wang, Zerong Qi, Shujun Fu$^*$ \\
  School of Matahematics, Shandong University, Jinan 250100, China \\
  \texttt{\{Shujun Fu\}\quad shujunfu@163.com} \\
   \And
  Mingzheng Hu \\
  Shandong Chengshi Electronic Technology Limited Company, Jinan 250002, China \\
}

\begin{document}
\maketitle

\begin{abstract}  
Latent fingerprint enhancement is a critical step in the process of latent fingerprint identification. Existing deep learning-based enhancement methods still fall short of practical application requirements, particularly in restoring low-quality fingerprint regions. Recognizing that different regions of latent fingerprints require distinct enhancement strategies, we propose a Triple Branch Spatial Fusion Network (TBSFNet), which simultaneously enhances different regions of the image using tailored strategies. Furthermore, to improve the generalization capability of the network, we integrate orientation field and minutiae-related modules into TBSFNet and introduce a Multi-Level Feature Guidance Network (MLFGNet). Experimental results on the MOLF and MUST datasets demonstrate that MLFGNet outperforms existing enhancement algorithms.
\end{abstract}


\section{Introduction}
Given the permanence and uniqueness of fingerprints, fingerprint recognition has been widely applied as a biometric technology across various domains. Latent fingerprints, which are the impressions left on surfaces after finger contact, serve as critical evidence in forensic investigations. Researchers extract and analyze latent fingerprints to determine the identity of suspects. However, due to typically small area, high noise levels, and severe local degradation, traditional fingerprint identification algorithms struggle to accurately identify latent fingerprints. During the identification process, forensic experts often manually annotate identifiable features to improve identification accuracy.

To improve the quality of latent fingerprints and reduce the reliance on manual intervention, latent fingerprint enhancement has become a crucial step in the identification process. In the early stages of latent fingerprint enhancement research, efforts were primarily focused on accurately extracting fingerprint orientation fields and designing corresponding filters to eliminate noise and ridge distortions. However, fixed-form filters often fail to achieve the desired enhancement results. With the success of deep learning in various computer vision tasks, researchers have proposed numerous deep learning-based latent fingerprint enhancement algorithms. While these methods integrate deep learning techniques with domain-specific fingerprint knowledge and have achieved promising enhancement results, they still fall short of practical application requirements, particularly in the restoration of low-quality fingerprint regions.

This study addresses the varying enhancement requirements of different fingerprint regions—high-quality regions, low-quality regions, and background regions—by designing a latent fingerprint enhancement network, TBSFNet, that applies distinct enhancement strategies to different regions. Furthermore, by incorporating fingerprint-related feature modules, we propose an improved network, MLFGNet, aiming to achieve superior enhancement performance across various low-quality latent fingerprint images.

\section{Proposed Methods}

\subsection{TBSFNet}

Latent fingerprint images can be divided into three main regions: high-quality fingerprint regions, low-quality fingerprint regions, and background regions. To effectively enhance these different regions, fingerprint enhancement algorithms must adopt distinct strategies: In high-quality regions, the enhancement algorithm should improve the contrast between ridges and valleys to facilitate subsequent minutiae extraction; in low-quality regions, the algorithm should leverage surrounding information to restore the structural integrity of fingerprint ridges; and in background regions, the algorithm should suppress structural noise that may interfere with identification results.

To achieve these objectives, we propose a novel Triple-Branch Spatial Fusion Network (TBSFNet), which consists of three independent branches dedicated to processing different regions of the fingerprint image. When applying TBSFNet for latent fingerprint enhancement, the weight maps generated by the three branches correspond approximately to the high-quality, low-quality, and background regions of the fingerprint image. This enables the simultaneous enhancement of different fingerprint regions within a unified network architecture.

\begin{figure}
	\centering
	\includegraphics[width=0.8\textwidth]{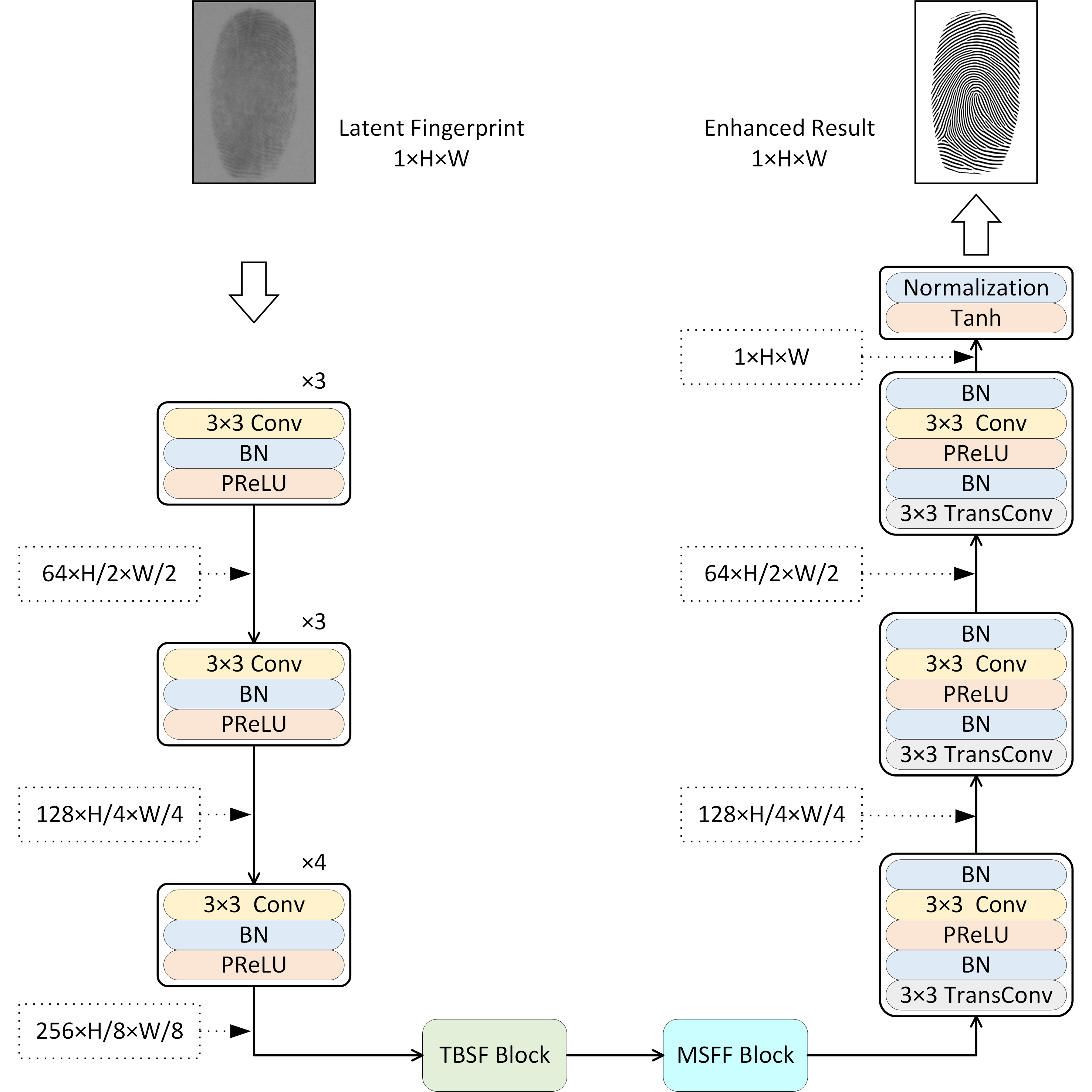}
	\caption{\centering Architecture of proposed TBSFNet  network. The input is a grayscale image with height H and width W.}
	\label{fig:1}
\end{figure}

The framework of the proposed TBSFNet is illustrated in Figure \ref{fig:1}. It adopts an encoder-decoder structure as its fundamental framework, leveraging the compression in the encoding phase and the reconstruction in the decoding phase to effectively remove noise, making it well-suited for latent fingerprint enhancement tasks.

In the encoding phase, the network first performs three downsampling operations using a combination of 3×3 convolutions, batch normalization, and PReLU activation functions. This process expands the input data channels from 1 to 256 while reducing the spatial dimensions to $\frac18$ of the original size. Subsequently, the Triple-Branch Spatial Fusion (TBSF) block is employed to specifically process different fingerprint regions. Additionally, the Multi-Scale Feature Fusion (MSFF) block, proposed by Shui et al.\cite{shui2024three}, is integrated to extract and merge multi-scale features effectively.

\begin{figure}
	\centering
	\subfloat[TBSF Block]{\includegraphics[width=0.8\textwidth]{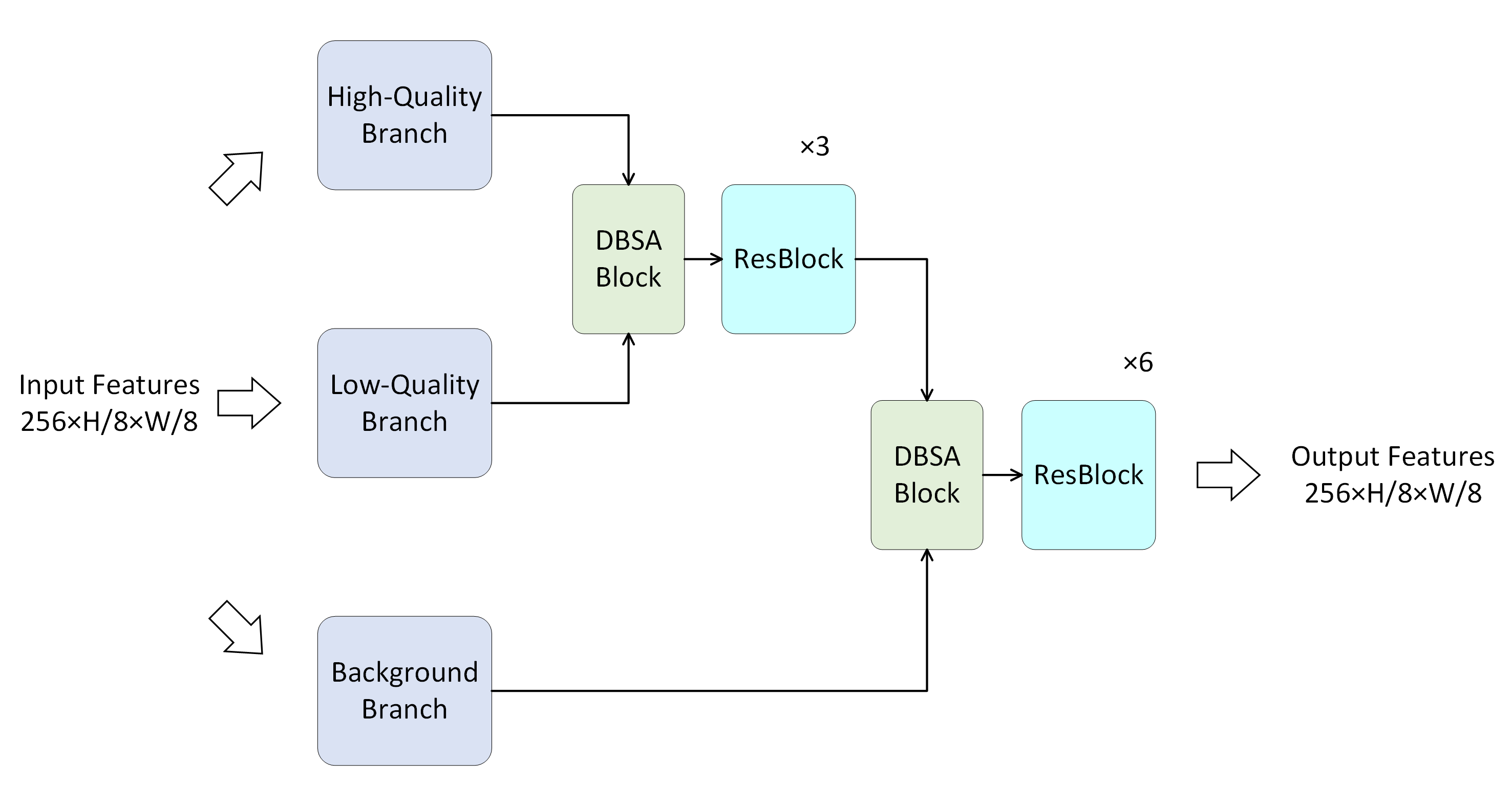}}
	\quad
	\centering
	\subfloat[DBSA Block]{\includegraphics[width=0.8\textwidth]{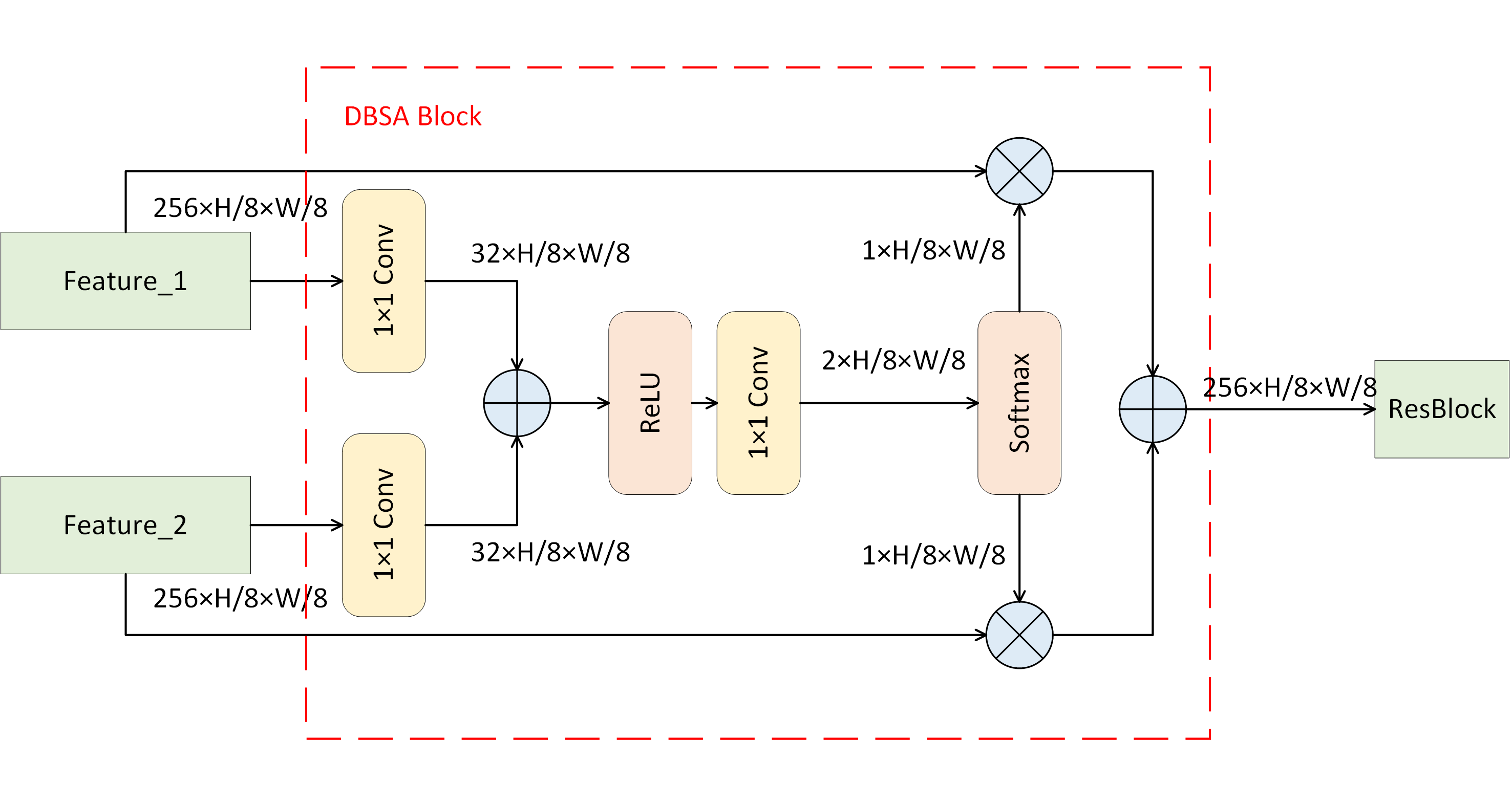}}
	\caption{Framework of TBSF and DBSA blocks.}
	\label{fig:2}
\end{figure}

As illustrated in Figure \ref{fig:2}(a), the TBSF block utilizes three structurally distinct branches to extract different features, followed by two stages of Double Branch Spatial Attention (DBSA) blocks to fuse the extracted features. Among these branches, the high-quality branch consists of three residual blocks\cite{he2016deep}, while the low-quality and background branches incorporate two downsampling operations before the residual blocks and two upsampling operations afterward to expand the receptive field of the module.The TBSA block is designed based on the well-known spatial attention mechanism Attention Gate (AG)\cite{oktay2018attention}. Its detailed structure is depicted in Figure \ref{fig:2}(b).

In the decoding phase, the network performs three upsampling operations using a combination of 3×3 transposed convolutions, batch normalization, PReLU activation functions, and 3×3 convolutions. This sequence effectively upsamples the data. Finally, the data is mapped to the range $[0, 1]$ using a Tanh activation function and linear normalization, resulting in an enhanced fingerprint image with the same spatial dimensions as the input.

\subsection{MLFGNet}

\begin{figure}
	\centering
	\includegraphics[width=0.8\textwidth]{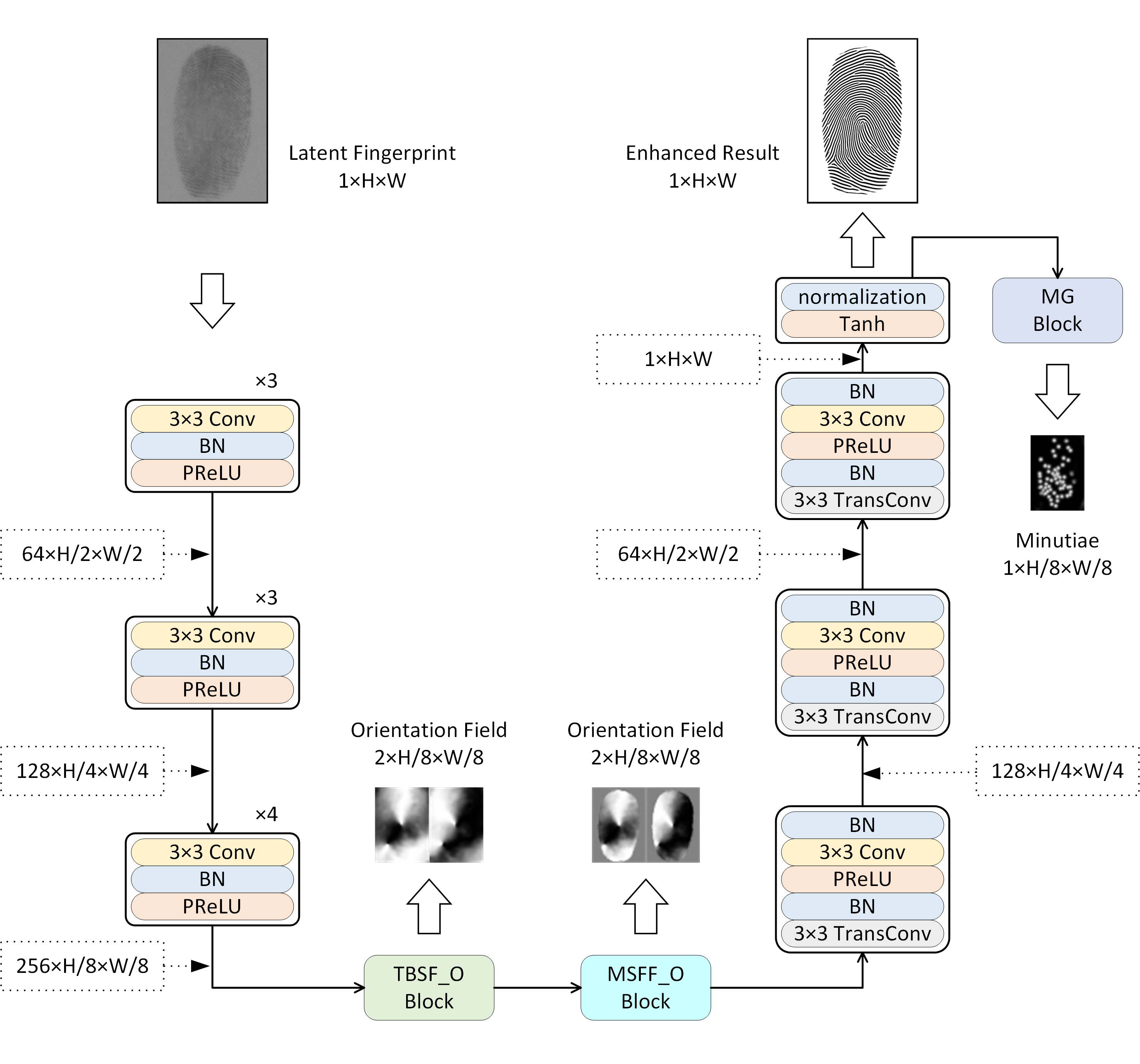}
	\caption{\centering Framework of MLFGNet network.}
	\label{fig:3}
\end{figure}

Building upon TBSFNet, we introduced orientation field and minutiae-related modules to propose the Multi-Level Feature Guidance Network (MLFGNet), as shown in Figure \ref{fig:3}. First, the MSFF\_O block estimates the orientation field using multi-scale features, guiding the network to extract ridge-direction-related features. Simultaneously, the TBSF\_O block outputs the orientation field directly after the low-quality branch to enhance the branch's ability to capture global features. Finally, after the enhanced output of TBSFNet, the Minutiae Guidance (MG) module is added to estimate minutiae regions while increasing the model's focus on minutiae areas.

\begin{figure}
	\centering
	\subfloat[MSFE Block and MSFS Block]{\includegraphics[width=0.8\textwidth]{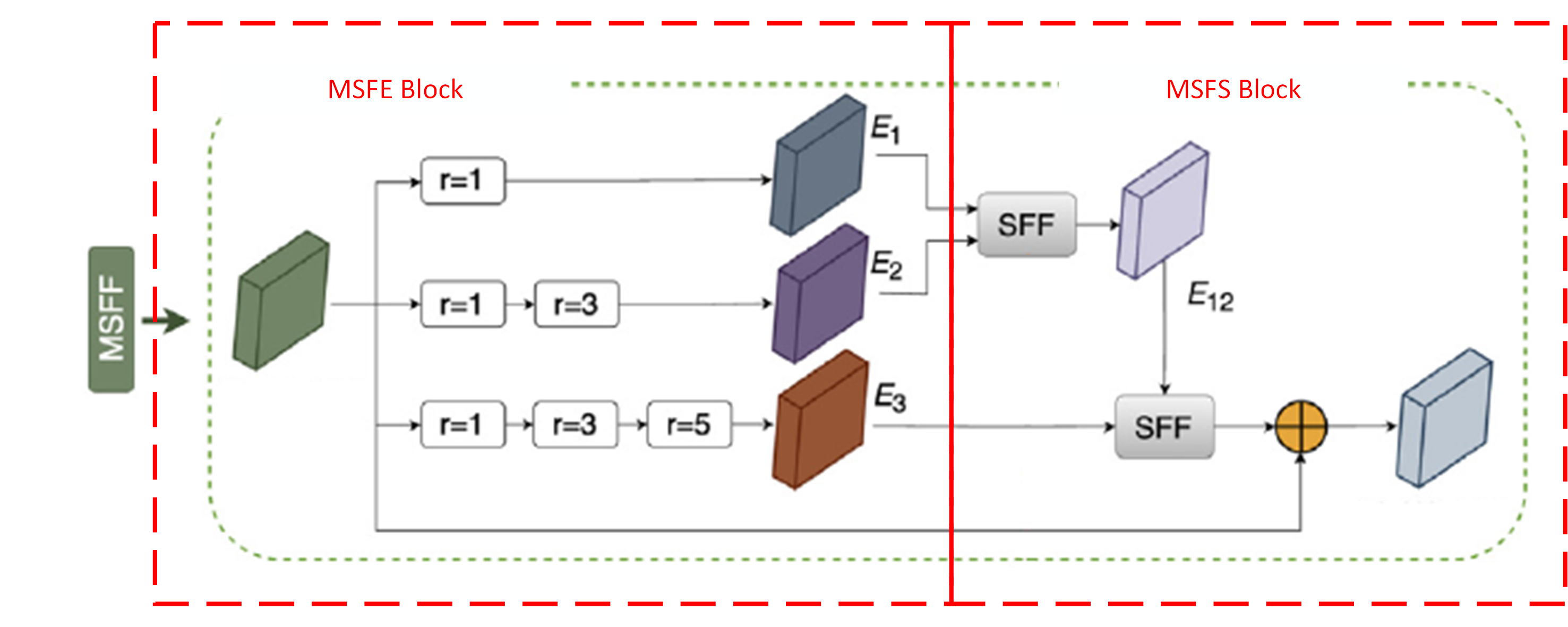}}
	\quad
	\centering
	\subfloat[MSFF\_O Block]{\includegraphics[width=0.6\textwidth]{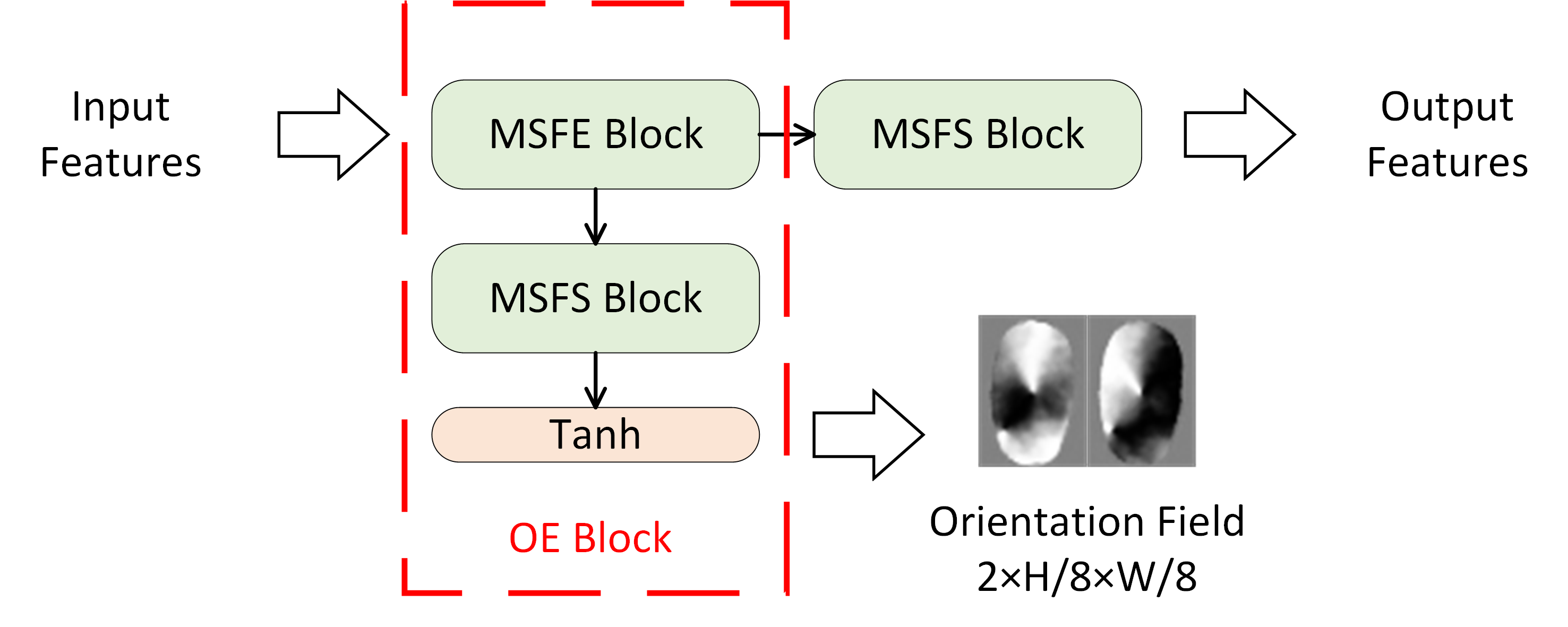}}
	\caption{Framework of MSFF\_O block.}
	\label{fig:4}
\end{figure}

Considering the different feature scales used for orientation field estimation and enhancement tasks, we decompose the MSFF module into two components: Multi-Scale Feature Extraction (MSFE) and Multi-Scale Feature Selection (MSFS) blocks, as shown in Figure \ref{fig:4}(a). Additionally, after the MSFE block, a MSFS block with two output channels (corresponding to the sine and cosine of the ridge direction at twice the angle) is added, along with a Tanh function, to estimate the orientation field, as shown in Figure \ref{fig:4}(b). For clarity, the structure composed of the MSFE block, the MSFS block with two output channels, and the Tanh function is referred to as the Orientation Estimation (OE) block.

\begin{figure}
	\centering
	\includegraphics[width=\textwidth]{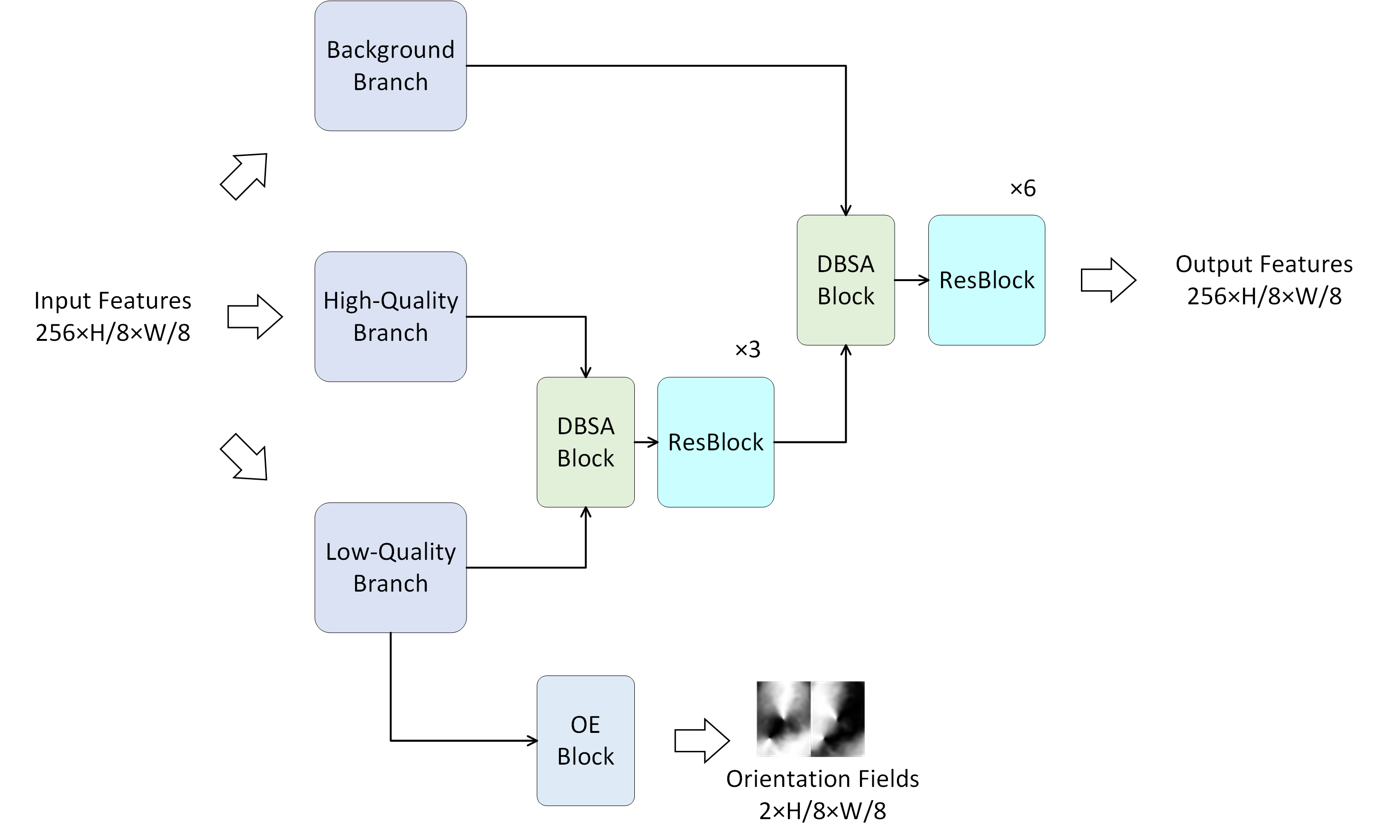}
	\caption{\centering Framework of TBSF\_O block.}
	\label{fig:5}
\end{figure}

Additionally, by adding the OE block after the low-quality branch, we propose the TBSF\_O Block (as shown in Figure \ref{fig:5}). Compared to the high-quality and background regions, the restoration of low-quality regions requires a larger neighborhood context. Therefore, the low-quality branch employs more downsampling operations and residual blocks to expand the receptive field of the branch. Directly estimating the global feature orientation field after the branch further improves the ability of the low-quality branch to integrate global information.

Recognition is the final goal of fingerprint-related tasks, and the design of fingerprint enhancement algorithms needs to consider the matching process. Minutiae-based matching methods are currently the predominant identification technique, and many enhancement networks utilize minutiae information to guide network training. In 2017, FingerNet\cite{tang2017fingernet} introduced a minutiae estimation module after the enhancement process to predict the locations and orientations of minutiae in the image. In 2023, FingerGAN\cite{zhu2023fingergan} used the skeleton map, which directly reflects minutiae information, as the network output, and increased the reconstruction loss weight in minutiae regions to enhance the network's focus on minutiae. In 2024, FingerRT\cite{jia2024finger} designed a Transformer-based model for fingerprint restoration, incorporating the number of minutiae as prior knowledge into the self-attention module.

\begin{figure}
	\centering
	\includegraphics[width=0.8\textwidth]{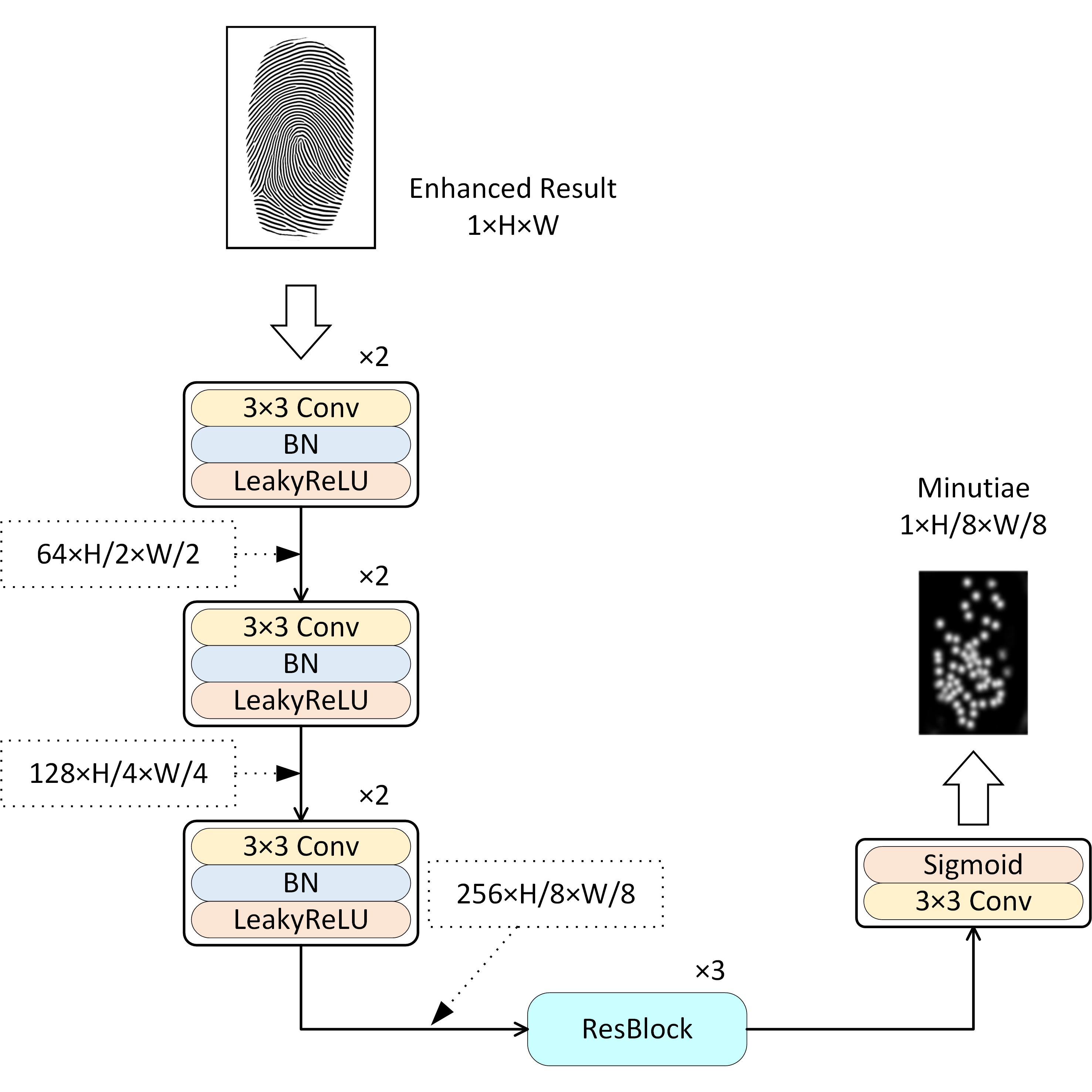}
	\caption{\centering Framework of MG block.}
	\label{fig:6}
\end{figure}

Inspired by FingerNet and FingerGAN, we add the MG block after the enhancement network, as shown in Figure \ref{fig:6}. The MG block first performs three downsampling operations using multiple 3×3 convolutions, batch normalization, and LeakyReLU structures, transforming the input data from 1 channel to 256 channels while reducing the size to $\frac18$ of the original. It then uses three residual blocks to further extract features and finally outputs the predicted minutiae regions through a 3×3 convolution followed by a Sigmoid function.

During the training process, the MG block is trained separately from the main network. After training the MG block using the ground truth of enhancement and corresponding minutiae regions, it is then used to predict the minutiae regions in the main network’s enhanced output. The difference between the predicted and actual minutiae regions is used to guide the network's parameter updates. Additionally, to further increase the influence of minutiae regions on the network, the reconstruction loss weight in the minutiae regions is amplified based on the predicted regions.

\section{Experiment Results}

\begin{figure}[htbp]
	\centering
	\subfloat[\centering Latent Fingerprint (MUST)]{\includegraphics[width=0.24\textwidth]{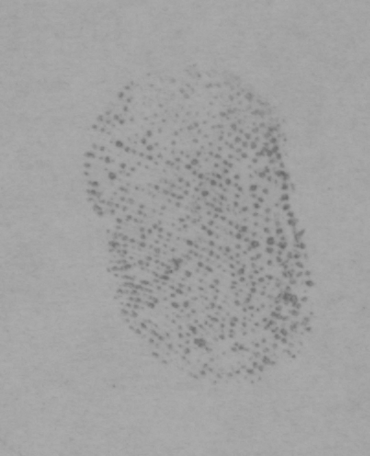}}
	\subfloat[\centering FingerNet]{\includegraphics[width=0.24\textwidth]{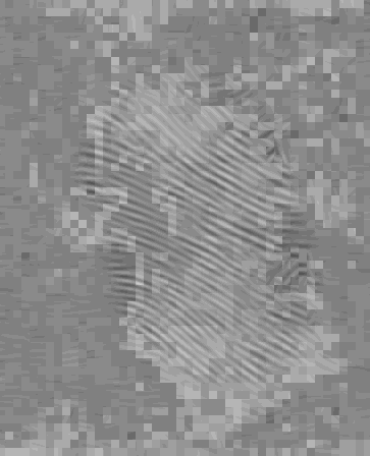}}
	\subfloat[\centering LatentAFIS]{\includegraphics[width=0.24\textwidth]{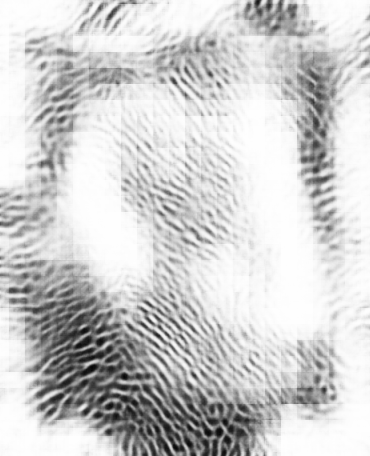}}
	\subfloat[\centering FingerGAN]{\includegraphics[width=0.24\textwidth]{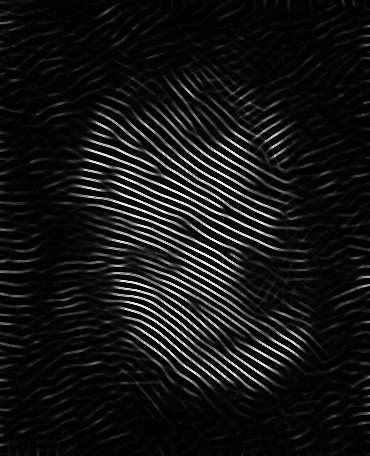}}
	\quad
	\subfloat[\centering JoshiGAN]{\includegraphics[width=0.24\textwidth]{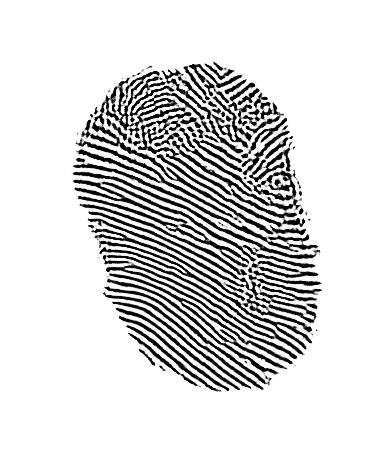}}	
	\subfloat[\centering COOGAN]{\includegraphics[width=0.24\textwidth]{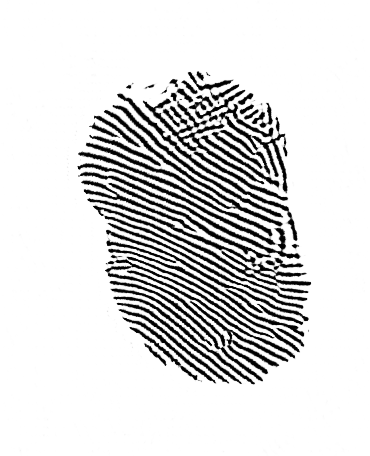}}
	\subfloat[\centering OFFIENet]{\includegraphics[width=0.24\textwidth]{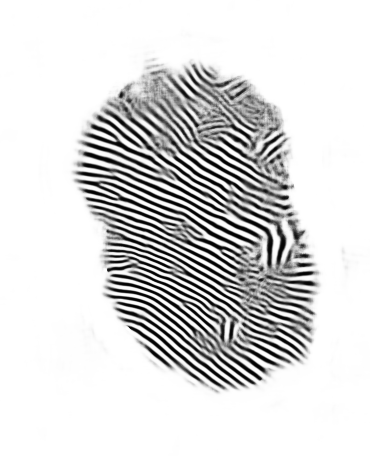}}
	\subfloat[\centering NestedUNets]{\includegraphics[width=0.24\textwidth]{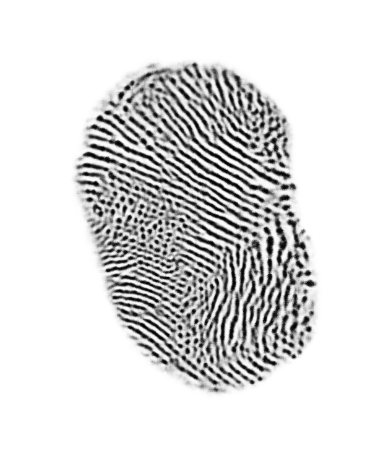}}	
	\quad
	\subfloat[\centering TBSFNet]{\includegraphics[width=0.24\textwidth]{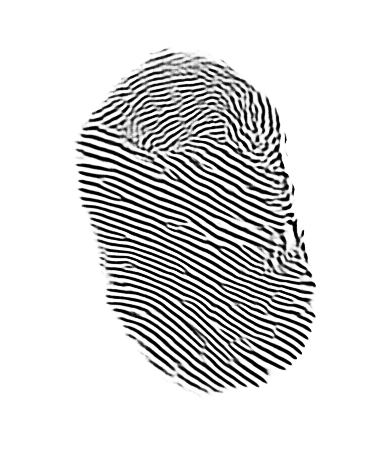}}
	\subfloat[\centering MLFGNet]{\includegraphics[width=0.24\textwidth]{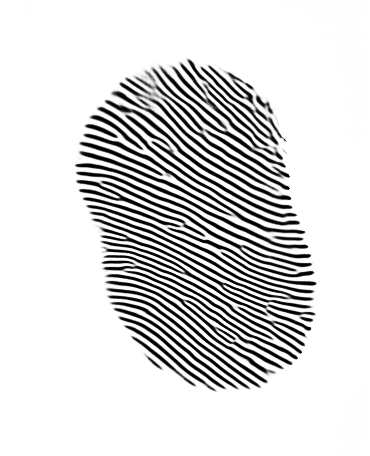}}
	\caption{\centering Enhancement results of some latent fingerprint enhancement networks on a latent fingerprint in the MUST dataset.}
	\label{fig:7}
\end{figure}

To demonstrate the superiority of TBSFNet and MLFGNet, we compare their enhancement results with existing enhancement algorithms (FingerNet\cite{tang2017fingernet}, LatentAFIS\cite{cao2019end}, FingerGAN\cite{zhu2023fingergan}, JoshiGAN\cite{joshi2019latent}, COOGAN\cite{liu2019cooperative}, OFFIENet\cite{wong2020multi}, NestedUNets\cite{liu2020automatic}). Figure \ref{fig:7} presents the enhancement results of these reproduced enhancement algorithms, along with the proposed TBSFNet and MLFGNet, on a fingerprint from the MUST dataset.

In this sample, due to the dryness of the volunteer's finger during pressing, the latent fingerprint ridges appear as discrete dots, making minutiae extraction more challenging and requiring enhancement. In some regions, such as the upper part of the fingerprint, the ridge points are highly dispersed. Algorithms that do not consider the orientation field often generate large areas where ridge directions change abruptly (e.g., Figure \ref{fig:7}(c), (e), and (h)). In contrast, algorithms that incorporate the orientation field (e.g., Figure \ref{fig:7}(b), (d), (f), and (g)) produce overall smoother ridge structures but still exhibit small regions where ridges break suddenly or change direction sharply.

Since TBSFNet does not incorporate the orientation field, its enhancement results still contain regions where ridge directions shift abruptly. However, MLFGNet, which extends TBSFNet with two OE blocks and an MG block, corrects the erroneous enhancements in the upper fingerprint region. As a result, MLFGNet produces more natural ridge structures and achieves the best visual enhancement quality among all compared algorithms.

While visual results intuitively demonstrate the impact of the OE block on the network, they do not directly highlight the contribution of the MG block. Since the MG module is designed to improve the identification accuracy, its effectiveness needs to be validated through fingerprint matching experiments.

Table \ref{tab:1} presents the Rank-1 and Rank-20 identification accuracy on the MUST dataset, as well as the Rank-1 and Rank-50 identification accuracy on the MOLF dataset for various algorithms. TBSFNet outperforms existing enhancement algorithms in terms of identification accuracy on the MOLF dataset but achieves results similar to FingerGAN on the MUST dataset. After integrating the OE and MG blocks, MLFGNet achieves significant improvements in all identification accuracy. Specifically, on the MUST dataset, MLFGNet's Rank-1 and Rank-20 identification accuracy increase by 0.0143 and 0.0193, respectively, compared to TBSFNet, surpassing FingerGAN in Rank-1 accuracy. Additionally, on the MOLF dataset, MLFGNet improves the Rank-1 and Rank-50 recognition rates by 0.0189 and 0.0272, respectively, compared to TBSFNet. Furthermore, MLFGNet outperforms OFFIENet, the best-performing existing algorithm, by 0.0669 and 0.0961 in Rank-1 and Rank-50 accuracy, respectively.

\begin{table}
	\centering
	\caption{Rank-N identification accuracy of latent fingerprint enhancement algorithms on MUST and MOLF datasets.}
	\begin{tabular}{c|cc|cc}
		\toprule
		   & \multicolumn{2}{c|}{MUST dataset} & \multicolumn{2}{c}{MOLF dataset} \\
		 & Rank-1 & Rank-20 & Rank-1 & Rank-50 \\
		\hline
		raw image   & 0.1083 & 0.1487 & 0.005 & 0.0211 \\
		FingerNet & 0.1366 & 0.2148 & 0.1256 & 0.2672 \\
		LatentAFIS & 0.1558 & 0.2406 & 0.2139 & 0.4306 \\
		FingerGAN & 0.2167 & 0.3034 & 0.2553 & 0.4594 \\
		JoshiGAN & 0.1863 & 0.2672 & 0.3156 & 0.5286 \\
		COOGAN & 0.1937 & 0.2907 & 0.305 & 0.5114 \\
		OFFIENet & 0.2032 & 0.2898 & 0.3206 & 0.5317 \\
		NestedUNets & 0.1961 & 0.2735 & 0.2847 & 0.4856 \\
		TBSFNet & 0.2156 & 0.3099 & 0.3686 & 0.6006 \\
		\textbf{MLFGNet} & \textbf{0.2299} & \textbf{0.3292} & \textbf{0.3875} & \textbf{0.6278} \\
		\bottomrule
	\end{tabular}
	\label{tab:1}
\end{table}%

In summary, MLFGNet achieves outstanding performance in both visual enhancement quality and identificaition accuracy, demonstrating a significant advantage over existing algorithms.

\section{Conclusion}
Recognizing that different fingerprint regions have distinct enhancement objectives, we designed TBSFNet, a triple-branch latent fingerprint enhancement network. By employing a spatial attention mechanism to fuse the features extracted from the three branches, TBSFNet effectively applies different enhancement strategies to different fingerprint regions.

Furthermore, fingerprint features play a crucial role in traditional fingerprint-related tasks, leading many deep learning-based fingerprint algorithms to integrate fingerprint-specific features into their networks. We observed that the global feature—orientation field—can guide fingerprint reconstruction, particularly in low-quality regions, while estimating minutiae regions in the enhancement output improves the network’s focus on local minutiae details. Based on these insights, we extended TBSFNet by incorporating orientation field and minutiae-related modules, leading to the development of MLFGNet. 

Experimental results demonstrate that MLFGNet outperforms TBSFNet on the MOLF and MUST datasets, exhibiting a clear advantage over existing enhancement algorithms.

\section*{Acknowledgments}
The research has been supported in part by the National Natural Science Foundation of China (12071263, 11971269, 12171285, 12371492), the Young Taishan Scholars Program (tsqn202211321), the Distinguished Taishan Scholars Program (tstp20231251), and the innovation ability improvement project of science and technology-based SMEs in Shandong Province (NO. 2022TSGC2072).

\bibliographystyle{unsrt}  
\bibliography{references}

\end{document}